\documentclass{ecai} 

\usepackage{microtype}
\usepackage{graphicx}
\usepackage{subfigure}
\usepackage{booktabs}

\usepackage{latexsym}
\usepackage{amssymb}
\usepackage{amsmath}
\usepackage{amsthm}
\usepackage{booktabs}
\usepackage{enumitem}
\usepackage{graphicx}
\usepackage{color}

\usepackage{url}            
\usepackage{amsfonts}       
\usepackage{nicefrac}       
\usepackage{algorithmic}
\usepackage{algorithm2e}

\newcommand{\BibTeX}{B\kern-.05em{\sc i\kern-.025em b}\kern-.08em\TeX}

\begin{document}

\begin{frontmatter}

\paperid{2657}

\title{Reducing Reward Dependence in RL Through Adaptive Confidence Discounting}

\author[A]{\fnms{Muhammed Yusuf}~\snm{Satici}\thanks{Corresponding Author. Email: msatici@ncsu.edu.}}
\author[B]{\fnms{David L.}~\snm{Roberts}}

\address[A]{Department of Computer Science, NCSU}
\address[B]{Department of Computer Science, NCSU}

\begin{abstract}
In human-in-the-loop reinforcement learning or environments where calculating a reward is expensive, the costly rewards can make learning efficiency challenging to achieve. The cost of obtaining feedback from humans or calculating expensive rewards means algorithms receiving feedback at every step of long training sessions may be infeasible, which may limit agents' abilities to efficiently improve performance. Our aim is to reduce the reliance of learning agents on humans or expensive rewards, improving the efficiency of learning while maintaining the quality of the learned policy. We offer a novel reinforcement learning algorithm that requests a reward only when its knowledge of the value of actions in an environment state is low. Our approach uses a reward function model as a proxy for human-delivered or expensive rewards when confidence is high, and asks for those explicit rewards only when there is low confidence in the model's predicted rewards and/or action selection. By reducing dependence on the expensive-to-obtain rewards, we are able to learn efficiently in settings where the logistics or expense of obtaining rewards may otherwise prohibit it. In our experiments our approach obtains comparable performance to a baseline in terms of return and number of episodes required to learn, but achieves that performance with as few as 20\% of the rewards. 

\end{abstract}

\end{frontmatter}

\section{Introduction}

Having the ability to learn from sparse rewards is a crucial aspect of real-world applications of reinforcement learning (RL) since, as opposed to the proxy reward functions defined in computer simulations, the real world events do not always return immediate and accurate feedback signals to the learner. These sparse rewards present challenges to the RL agents due to their irregular and imperfect distribution. Many sparse-reward RL algorithms introduce artificial rewards to densify the reward function in the hope of facilitating the learning process and overcoming the sporadic nature of the reward signal. We take the opposite approach in this research. We assume the agent has the ability to request reward signals at will from the environment. We start with relatively dense reward signals at the beginning of the training and make the reward signals sparser in line with the improvements in the agent's learning model. In this way, we prevent the agent from receiving rewards in the parts of the state space where the agent already has a good understanding of its policy and gather reward signals only for the states that the agent has a low confidence in its action selection. The agent depends on the external feedback at the early stages of its learning but does not require frequent feedback signals once it obtains a certain amount and variety of experiences.

Our algorithm, at a high level, measures the learner's confidence 
using the output distribution(s) of the agent's model(s).
We assume the agent has the ability to request feedback from the environment for the given state and train a reward function model using the results of those requests. The agent is trained using regular deep RL algorithms, but for the states with no requested feedback value, the agent uses the output of the reward function model as the feedback. In this way, the agent skips the feedback from the environment for the states it has high confidence.

We compare two different formulations for measuring confidence. The first uses the entropy derived from the output probability distribution of the actor model (or the Q-value distribution for DQN architectures)---low entropy suggests high confidence in action selection. The second uses entropy derived from the output distribution of the reward function model in addition to the entropy of the actor model. This combination of reward and action entropies measures the inaccuracies in both the agent's learning and learning the reward function. We also compare two different regularization terms to lessen the reduction in environment rewards. We test all of our approaches in three domains: a discrete-space sparse-reward grid-world, a continuous-space sparse-reward robotics environment, and a continuous-space dense-reward highway environment. We evaluate our approach against deep Q-network (DQN)~\cite{humanLevelNN}, actor-critic network (A2C)~\cite{mnih2016asynchronous}, and hindsight experience replay (HER)~\cite{2018hindsight}. Results show that our approach at worst matches the cumulative reward of the baselines while greatly reducing the number of environment rewards. 

\section{Related Work}

One challenge in applying RL to real-world problems is the cost of sample collection. Sample efficient RL algorithms have been proposed in recent years to lower the sample complexity of deep RL algorithms in real world scenarios \citep{NEURIPS2018_f02208a0, mai2022sample, zhang2020sample}. These approaches aim to mitigate the challenges posed by the high costs associated with collecting real-world samples for applications of RL. In the case of our algorithm, instead of focusing on the cost of interacting with the environment, we investigate the problem of improving the feedback efficiency by making the training process as independent as possible from the rewards returned by the environment. The feedback efficiency differs from the sample efficiency in that feedback efficient methods afford requesting and training on many transitions from the environment as opposed to sample efficient algorithms but they cannot afford asking guidance from a human expert due to the high cost of feedback retrieval associated with it. There have been few deep-RL algorithms aiming to learn with limited feedback derived from human preferences \citep{kong2023provably, lee2021pebble, park2022surf} but none of those approaches offer a framework that does not depend on human feedback elicitation. We formulate the feedback retrieval problem as an implicit curriculum and try to offer a general framework that works on both human-designed and artificial feedback.

Algorithms that deal with sparse rewards focus on encouraging exploration to increase the agent's chance of attaining useful reward signals. One common approach defines a curiosity function to measure the novelty of the visited states and provides the agent additional reward signals if it visits a novel state of the environment~\cite{curdriven}. Another approach uses the error of the neural network as an exploration bonus added to the reward function to encourage the agent to visit the areas of the state space with high error in the prediction of the observation features~\cite{networkdistil}. A different approach employs a HER buffer to sample additional transitions containing the future state of the agent as the goal for the transition at hand~\cite{2018hindsight}. These methods try to address the sparsity of rewards as an issue of limited exploration and aims to improve the agent's performance by bringing it closer to unknown regions of state space. They do not change the number of rewards, but try to put the agent in a position where it could gather the most novel feedback. Our algorithm does not encourage exploration; rather, it prevents the agent from receiving unnecessary rewards.

There are also human-guided RL algorithms that attempt to model the human feedback using a reward predictor and use this reward model to train the agent without having the need to design a proxy reward function. These algorithms are typically used for complex real-world environments where it would be difficult to craft a reward function that encapsulates all aspects of the RL objective. Deep TAMER learns a reward function from human-guided feedback and uses this reward function in training the behavioral policy of the agent~\cite{warnell2018deep}. Similarly, Christiano {\em et al}.\ trains a reward function predictor using human-guided and synthetic feedback~\cite{christiano2023deep}.
Liang {\em et al}. uses disagreement between reward functions to increase exploration in human preference-based rl algorithms \cite{liang2022reward} and Liu {\em et al}. optimizes the reward model, and the agent in parallel in a bi-objective optimization process rather than learning a reward model before the agent training, which is the case for the majority of the human-guided RL algorithms \cite{metaReward}. 
These approaches show that it is possible to derive a useful reward function model through human feedback elicitation but they make no attempt at reducing the agent's dependency on the feedback coming from the environment. They also only depend on the feedback from the reward predictor in training their agent. On the other hand, we use both the reward from the environment and the feedback from the reward function model in training and show it is possible to dramatically reduce the environment rewards used during training. 

Finally, there are inverse RL (IRL) algorithms that build a model of the reward function from demonstrations. These algorithms share common elements with the sparse-reward RL in that they both aim to obtain a reward predictor. However, IRL algorithms address the problem of inferring a good reward function from demonstrations whereas sparse-reward RL uses a reward model for making the reward function less sparse or more efficient. Although IRL algorithms could be used to deal with sparse-reward learning, they cannot attain that goal without re-purposing their inferred reward model. 
See Arora {\em et al}.~\cite{irl} for a detailed analysis of IRL.

\section{Problem Formulation}

Here we provide background on the RL method and formally define the feedback diminution problem.

\subsection{Reinforcement Learning}\label{rl}

We model learning as a MDP \(M \) \(=\) \(<S\), \(A\), \(T\),
\(R\), \(s_i\), \(S_g>\), a tuple consisting of a set of states \(S\), a set of actions
\(A\), a transition function \(T\), a reward function \(R\), an initial state \(s_i\),
and a set of terminal states \(S_g\). The transition function \(T:S \times A
\times S \rightarrow [0,1]\) corresponds to the probability of transitioning
from a state \(s \in S\) to another state \(s' \in S\) using a valid action \(a \in
A\) at state \(s\). All of the environments we use are deterministic, so taking
action \(a\) in state \(s\) always transitions into the same resulting state
\(s'\)---although that is not a requirement for our algorithm. The reward function
\(R:S \times A \times S \rightarrow F\) maps a state, action, state tuple to
the real-valued reward the learner receives. The
policy \(\pi: S \rightarrow A\) maps states to actions. The cumulative reward \(G\)
at time \(t\) is the discounted sum of all rewards the agent receives from \(t\)
until it reaches a terminal state, 
    $G_t = \sum_{k=0}^{\tau-t} \delta^{k}*R_{t+k}$,
where \(\delta\) is the discount factor, \(\tau\) is the time to reach a
terminal state, and \(R_{t+k}\) is the feedback received in state
\(s_{t+k}\)~\cite{SuttonRL}. The agent's objective is to learn the optimal policy
\(\pi^*_M\) that maximizes \(G\).

We train using deep Q-networks (DQN)~\cite{humanLevelNN} for discrete space environments. We use two
four-layered, fully-connected, feed-forward DQNs. The networks take the
states as input and output Q-value estimates for each available action. One
neural network serves as the learned model, and the other provides target Q-value
estimates. The loss function is
\begin{equation}\label{eq1}
    \begin{split}
    L(\theta) = E_{s, a, r, trm, s' \sim RB} [ (r + \delta * & max_{a'} Q(s', a'; \theta^-) \\ & - Q(s, a; \theta))^2 ],
    \end{split}
\end{equation}
where \(\theta\) is the learned model weights, \(\theta^-\) is the
target model weights, and \(\delta\) is the discount
factor~\cite{humanLevelNN}. 

For continuous action spaces, we employ an actor-critic architecture similar to~\cite{mnih2016asynchronous}. We use a
four-layered, fully-connected, feed-forward network for the actor and critic. The critic receives the state as its input and outputs the value function estimate. The actor network outputs two real vectors which we treat as the mean and standard deviation of the multi-dimensional normal distribution that we sample the actions from. We use the advantage loss defined in~\cite{mnih2016asynchronous} to train the actor, and we train the critic network using the mean square error given in Eq.~\ref{eq1}. In the case of the robotics environment that we discuss in the results section, in addition to the given actor-critic architecture, we also employ a hindsight experience replay buffer (HER) to up-sample the positive reinforcement transitions. We use the future strategy with $k=4$ for the HER algorithm since that strategy is shown to produce the best results in~\cite{2018hindsight}. We also use the same loss function as~\cite{2018hindsight} in the robotics environment while retaining the aforementioned actor network for the sampling of the action values. 

At each training step, the agent takes a single
action in the environment, records the tuple \((s, a, r, trm, s')\) into a
replay buffer \(RB\) (where \(trm\) indicates whether \(s\) is a terminal state),
randomly samples tuples from the RB, and performs a single batch
update on the learned model. The target model weights are updated using
the weights of the learned model at the end of each episode.

\subsection{Feedback Diminution Problem}

In environments where it is difficult to retrieve rewards but relatively easy to take actions and observe new states, there is a need for using feedback efficient RL algorithms. Real world applications of reinforcement learning such as disease outbreak simulations require extensive computations to assess the quality of an action being taken where allocation of resources to certain locations to mitigate the spread of the disease is necessary \cite{episim, sykes_estimating_2023}. In these scenarios, it becomes trivial to take an action in the environment and observe the change in the environment state for a single timestep but assessing the outcome of the action taken involving interactions between many locations possibly over the span of a long period of time proves time consuming. Feedback efficient RL agents show potential in addressing the shortcomings of designing good reward functions for these complex real world environments by calculating the reward only for certain states of the environment where the agent itself determines what states would require an associated reward value.

The feedback diminution problem asks how an agent could learn the optimal policy for a given task while requesting as little feedback as possible from the environment. Traditional RL agents receive a single feedback signal for every environment interaction. In the case of feedback diminution, the agent develops a model of the feedback function and uses this model to sparsify the feedback signals. The sparsification of the feedback is accomplished by allowing the agent to request a reinforcement only when it deems necessary.
By solely looking at its own understanding of the target task, whether that knowledge comes from the Q-values or the reward function model, the agent solves a bi-objective optimization problem that reduces the number of reward signals coming from the environment while maximizing the cumulative reward objective as is the case with the regular RL agent. 

We evaluate the performance of the RL agent based on the total reward it achieves with and without feedback diminution at a given time in the training (asymptotic performance). We also measure the performance based on the number of feedback signals the agent needs to converge to the maximum cumulative reward \(G^{max}\) (time to convergence). The second measure does not look at how much training the agent performs to converge since all of our algorithms perform the same amount of training; rather, it attempts to measure the cost of receiving rewards during the training process. Our experiments compare feedback diminution algorithms based on these two metrics. 

\section{Entropy Approaches for Feedback Diminution}

We present confidence measures, pseudo-code for the high-level description of our feedback diminution algorithm, and define the regularization terms for the entropy calculation. 

\subsection{Measuring Confidence}

We construct our confidence criteria using the output distributions of the actor and reward models. The probability distribution here could be defined over a discrete or continuous variable depending on how the neural networks treat output parameters. For the discrete case, we define the entropy as
\begin{equation}\label{eq2}
    H(\pi(\cdot | s)) = - \sum_{a \in A} P_1(s, a)*lg(P_1(s, a)),
\end{equation}
where \(H(\pi(\cdot | s))\) is the entropy of the action policy $\pi$ for the state \(s\) using the probabilities \(P_1\). We calculate \(P_1\) as
    $P_1(s, a) = \frac{e^{Q(s, a)}} {\sum_{a \in A} e^{Q(s, a)}}$,
for the DQN architecture where \(Q(s,a)\) is the Q-value for the state-action pair \((a,s)\), and \(P_1(s, a)\) is the softmax of
Q-values. For the continuous case,
we define the differential entropy as
\begin{equation}\label{eq3}
    H(R(s, a)) = - \int_{R(s, a)} P_2(R(s, a))*lg(P_2(R(s, a))*dR,
\end{equation}
where \(H(R(s,a))\) is the entropy of a single state, action pair w.r.t the reward function \(R\) modeled by a four-layered fully-connected neural network taking state and action as input and outputting the mean and standard deviation of the Gaussian distribution for probabilities \(P_2\). \(P_2\) differs from \(P_1\) in the sense that it does not use softmax to calculate probability values since the output of \(R\) is already a Gaussian distribution that we sample the probability values for \(P_2\). We use Eq.3 to calculate the reward entropy of the reward model \(R\) and use Gaussian negative log likelihood loss in reward model training. For actor models that output continuous actions instead of Q-value estimates, we use the entropy calculation given in Eq.~\ref{eq3} for the action entropy where R(s, a) is replaced by the output of the action policy $\pi$, which is again a Gaussian distribution described in Section \ref{rl}. Since differential entropy is not bound to a specific range as opposed to the discrete entropy, we clip the differential entropy values to [0,10] range and then normalize them to obtain a confidence measure that is at the same scale as the discrete entropy, which is always between [0,1]. In vast majority of the states, the differential entropy lies in [0, 10] so the clipping doesn't cause a significant change in the entropy calculation. We define the confidence of the agent as \(Conf(s) = 1 - H(R(s, a))\) for the reward model and \(Conf(s) = 1 - H(\pi(\cdot | s))\) for the action model. We take the harmonic mean of the two confidence values to obtain the final confidence value for state $s$.

\subsection{Feedback Diminution Algorithm}

Our algorithm calculates the entropy of the action and reward models for the current state, tells the environment whether it wishes to skip the reward based on the agent's confidence level, and trains the agent using the reward values sampled from the reward model if a transition does not have a feedback assigned to it. We use two reward models to improve the stability of the feedback prediction. One model serves as the learning model and the other the target. We use the target model to predict the missing reward values for the training of the agent. We train the learning model at every iteration and copy it to the target reward model after each episode. We keep two separate buffers: the replay buffer contains all transitions for training the agent and the feedback buffer stores only the transitions containing an associated reward value.
Algorithm~\ref{alg:algo1} is pseudo-code for the training process. 

    \begin{algorithm}[htb]
    \caption{High-level Algorithm}
    \label{alg:algo1}
    \begin{algorithmic}[1]
    \footnotesize
    \STATE \textbf{INPUTS:} environment: \(ENV\); convergence criteria: \(conv\).
    
    \STATE \textbf{CONSTANTS:} set of terminal states: \(S_g\); initial state: \(s_i\); \\ confidence threshold: \(CTHRESH\).
    
    \STATE \textbf{VARS:} replay buffer: \(RB\); feedback buffer: \(FB\); state: \(s\); \\ terminal condition: \(trm\); reward: \(r\). 
    
    \STATE Initialize agent model: \(NN_{agent}\);
    
    \STATE Initialize reward model: \(NN_{rew}^{learn}\) and \(NN_{rew}^{target}\)
    
    \WHILE{\(conv\) is not satisfied}
    
        \STATE \(s \gets s_i\) of \(ENV\) 
        
        \STATE \(trm \gets s \in S_g\)  
    
        \STATE \(r \gets null\)  
        
        \WHILE{\(trm\) is False}
        
        \STATE Select \(a\) using \(\epsilon\)-greedy policy on \(NN_{agent}(s)\)
    
        \STATE Calculate confidence \(Conf\) on \(NN_{agent}\) and \(NN_{rew}^{target}\)
        
        \STATE Take action \(a\) in \(ENV\) and observe \(s'\)
    
        \IF{\(Conf \leq CTHRESH\)}
            \STATE observe r from \(ENV\)
            \STATE Store \((s,a,r,trm,s')\) in \(FB\)
        \ENDIF
    
        \STATE Store \((s,a,r,trm,s')\) in \(RB\)
        
        \STATE Sample a minibatch B from \(FB\)
        \STATE Perform a batch update on \(NN_{rew}^{learn}\) using \(B\)
        
        \STATE Sample a minibatch B from \(RB\)
    
        \FOR{\((s,a,r,trm,s')\) in \(B\)}
        \IF{\(r == null\)} 
        \STATE \(r \gets NN_{rew}^{target}(s,a)\)
        \ENDIF
        \ENDFOR
         
        \STATE Perform a batch update on \(NN_{agent}\) using \(B\)
       
        \STATE \(s \gets s'\)
        \STATE \(trm \gets s \in S_g\)   
        \ENDWHILE    
    
        \STATE Copy \(NN_{rew}^{learn}\) to the \(NN_{rew}^{target}\)
        
    \ENDWHILE
    \end{algorithmic}
    \end{algorithm}

Algorithm~\ref{alg:algo1} first initializes the agent model (Line~4) which is the neural network we use for the policy, including the target network. Then, Algorithm~\ref{alg:algo1} initializes the reward models (Line~5) and begins training from an initial state of the environment (Line~7).
It performs \(\epsilon\)-greedy action selection from the agent model (Line~11) and observes the next state (Line~13). 
It calculates confidence using Eq.~\ref{eq2}~\&~\ref{eq3} (Line~12) and if confidence is below the threshold (Line~14), it receives a reward from the environment (Line~15). 
It stores $(s,a,r,trm,s')$ tuples in \(FB\) for the training of the reward model (Line~16) and in \(RB\) for the training of the agent (Line~18). Then, it samples a minibatch of transitions from \(FB\) (Line~19) and performs an optimization step on the learning reward model (Line~20). It also samples a minibatch of transitions from \(RB\) (Line~21) and replaces the missing reward values with the predictions from the target reward model (Line~24). Finally, it performs an optimization step on the agent using the minibatch from \(RB\) with the mixture of actual and predicted rewards values (Line~27). It repeats the training process until the agent reaches the end of episode (Line~29~\&Line~10). Then, it copies the learning reward model to the target reward model (Line~31) and continues until the convergence criteria is met (Line~6).

    \begin{algorithm}[htb]
    \caption{Confidence Calculation with Discrete Action and Continuous Reward Entropies}
    \label{alg:algo2}
    \begin{algorithmic}[1]
    \footnotesize
    \STATE \textbf{INPUTS:} learning model: \(NN_{agent}\); reward model: \(NN_{rew}^{target}\); \\ state: \(s\); action: \(a\).
    
    \STATE \textbf{VARS:} Q-values for state s: \(Q^s\); probabilities for agent and \\ reward models: \(P^{agent}\) \& \(P^{reward}\); entropies for agent and \\ reward: \(H(A)\) \& \(H(R)\). 
    
    \STATE \textbf{OUTPUTS:} agent's confidence for state s: \(Conf\).
        
    \STATE \(Q_{agent}^s \gets\) get Q-values from \(NN_{agent}(s)\)
    
    \STATE \(P_{agent} \gets softmax(Q_{agent}^s)\)
        
    \STATE \(P_{reward} \gets\) get \(\mu\), \(\sigma\) from  \(NN_{rew}^{target}(s, a)\)
    
    \STATE \(H(A) \gets entropy(P_{agent}^s)\)
        
    \STATE \(H(R) \gets entropy(P_{reward}^s)\)
    
    \STATE \(Conf \gets Harmonic Mean(1 - H(A), 1 - H(R))\)
    
    \STATE return \(Conf\)
    
    \end{algorithmic}
    \end{algorithm}

Algorithm~\ref{alg:algo2} provides the details for the entropy calculation given in Algorithm~\ref{alg:algo1} at Line~12. It takes the agent and reward models, and the current state-action pair as inputs. It takes the Q-values for the agent model (Line~4), applies softmax to obtain probability values (Line~5), and uses the probability estimates from the reward model (Line~6). It then calculates the action entropy (Line~7; H(A)), and the reward entropy (Line~8; H(R)), and converts them to confidence values (Line~9). It combines these confidence values using the harmonic mean (Line~9) since the harmonic mean tends to be closer to the lower of the two confidence values making the algorithm less greedy in skipping rewards. In our experiments, we refer to this entropy calculation as action entropy (AE) + reward entropy (RE) since it attempts to capture the confidence of the agent in both its ability to learn the reward function and its accuracy in the action policy.
We also use a simplified version of this algorithm where we only consider the entropy of the agent without any guidance from the reward model. We refer to the second version of the entropy calculation as action entropy (AE) since it does not use the harmonic mean to combine two entropy values.

\subsection{Regularization of Confidence}\label{reg}

The feedback diminution algorithm is a greedy process that skips the reward for every state where the agent's confidence is predicted to be high based on entropy calculations. During training, the agent does not always produce reliable estimates for its confidence since it has no access to the optimal policy and the entropy only serves as a heuristic assessment of confidence. Skipping the reward in states where the agent's confidence is misplaced may lead to suboptimal performance. To prevent this type of phenomena, we regularize the confidence of the agent. 

We offer two regularization terms, namely, exponential and hyperbolic regularization. We define exponential regularization as
    $e^{ - \nu * n}$,
and the hyperbolic regularization as
   $\frac{1}{1 + \nu * n}$,
where \(\nu\) is the temperature parameter and n is the number of steps taken without any reward from the environment ({\em i.e.}, we reset $n$ to zero when the agent receives a reward from the environment). 
We multiply the confidence term with the regularization term at every step to reduce the likelihood of the agent not receiving any reward for long periods of training. The exponential regularization performs a steeper reduction in the confidence values. Hyperbolic regularization provides a softer decay than exponential and resembles the human psychological decay for delayed rewards~\cite{e16105537}. We use a constant temperature parameter \(\nu\)  which is 0.5 for exponential and 1 for hyperbolic regularization. We did manual hyperparameter tuning in the range of [0.5, 0.75, 1] and picked the best performing value for the temperature parameter.  

\section{Evaluation Methods}

Here we describe the test environments, experimental setup, and state
representation for each domain.

\subsection{Environments}\label{test}

We use a $20 \times 20$ 2D grid and two continuous state-action space domains to test the algorithms. 

\subsubsection{Key-Lock}

The Key-Lock domain
contains keys, locks, pits, and obstacles similar to~\cite{KonidarisDomain}
and~\cite{NarvekarPolicyChange}.  The agent's task is to pick up the key and
unlock the lock while avoiding the pits and obstacles. Each key picked up gives
a reward of 500 and each lock unlocked gives a reward of 1,000. Falling into a
pit receives -400. All other actions including moving into an obstacle receive
-10.  Moving into an obstacle results in no state transition. The learner can
only move in cardinal directions and is assumed to have obtained the key or unlocked the
lock if its location matches the location of the key or it has obtained the key and matches the location of the lock. An episode
terminates when the agent obtains all the keys and unlocks all the locks, falls into a pit, or
reaches 100 time steps. Since the agent only receives rewards when it is exactly at the same position as the key or the lock, this is a sparse-reward task. A state in the key-lock environment is represented as a vector, including the Euclidean distance from the
learner in all cardinal directions to the nearest key and lock, four binary
parameters indicating if there is an obstacle in the neighboring cells, four binary parameters indicating if there is a key or lock in the neighboring cells, and eight binary parameters indicating if there is a pit in the two adjacent cells in all four directions. Lastly, two integers indicate the number of keys and locks captured so far. 

\subsubsection{Fetch-Push}

The Fetch-Push environment from the gymnasium robotics library~\cite{plappert2018multigoal} is a robotics control domain where a mobile manipulator robot must move a block to a target position using its gripper. The position of the robotic gripper and the block are randomly determined at the beginning of each episode. The agent takes three continuous actions that change the displacement of the gripper in 3D space. All of the actions are defined in the range of [-1, 1]. The reward the agent receives is 0 for having the block within the target position and -1 otherwise. Episodes don't terminate, instead they are truncated after 50 steps.
A state is a vector, including the position, velocity and displacement of the gripper and the position, velocity and rotation of the block. The input to the neural network is the concatenation of the state and the goal information since the environment uses a goal-aware observation space.   
\subsubsection{Parking}

The parking environment~\cite{highway-env} consists of 30 parking spots, an agent and a goal item randomly positioned in one of the spots. The agent always starts in the same position but its initial orientation changes randomly and is tasked to reach the goal and orient itself in the right direction. There are two continuous actions: velocity and angular velocity both in [-1,1]. An episode terminates when the agent reaches the goal and orients itself in the correct direction or when the episode length reaches 100 time steps. 
The environment allows the agent to wander outside of the parking lot and does not provide any boundaries. 
The agent receives a punishment proportional to its distance to the goal w.r.t. position and orientation.   
A state is a vector, including the agent's position, velocity, angular velocity, and the goal's position and orientation. The input to the neural network is the concatenation of the state and goal information since the environment uses a goal-aware observation space.

\subsection{Comparison Algorithm}

We compare our algorithm to a DQN~\cite{humanLevelNN} in the key-lock domain, a variant of advantage actor-critic network (A2C)~\cite{mnih2016asynchronous} in the parking environment and the hindsight experience replay (HER)~\cite{2018hindsight} algorithm in the robotics domain.
Our approach to reducing environment rewards makes the task harder for the agent to learn as time passes, which in return creates a soft curriculum without providing a clear sequence of curriculum tasks. HER also serves as an implicit curriculum learning algorithm as it makes reaching the goal simpler for the agent. However, our algorithm deals with reducing the feedback input to the agent whereas HER modifies the already existing transitions in the replay buffer. For this reason, we ran our approach on top of HER to get the benefit of both as they do not hinder or interfere with one another. 
For additional comparison, we also use a randomly generated entropy value between 0 and 1 as a random feedback diminution method and we use
 the regularization term without any entropy as a constant feedback diminution approach.

\subsection{Experimental Setup and Hyperparameters}

We compare performance based on the total reward obtained while learning the
task and the convergence times, and
include 95\% confidence intervals. All algorithms share the same hyperparameters for the neural network and train using the same environment-specific parameters.

\begin{table}[htb]
    \caption{Hyperparameters for the Feedback Diminution Algorithm.}
    \centering
    \footnotesize
     \begin{tabular}{ c c c c } 
        \toprule
    \textbf{Param} & \textbf{Value} & \textbf{Param} & \textbf{Value} \\ 
        \midrule
     \(\epsilon\) & 1 &  Optimizer & Adamax \\
    
     Eps. Decay & 0.995 & Loss & Mean Squared \\
    
     Min. \(\epsilon\) & 0.01 & Buffer Size & 40,000 \\ 
     
     \(\alpha\) & 0.005 & Batch Size & 16 \\
    
     \(\delta\) & 0.99 & \(\tau\) & 0.99 \\ 
    
    Conf. Thresh. & 0.25 & Averaging & Harmonic Mean \\ 
        \bottomrule
    \end{tabular}
    \label{table:hyper}
\end{table}

There are two types of hyperparameters: 1) neural network (including the
convergence criteria) and 2) entropy calculation. Table~\ref{table:hyper} contains the parameters used in our experiments.
These parameters are not
tuned for any specific algorithm and all algorithms we use share these
parameters for all of the experiments run in the same domain. The last row represents the parameters specific for the entropy calculation. All other parameters are for the neural network. \(\epsilon\) is the
\(\epsilon\)-greediness of the algorithm, which decays during training. \(\alpha\)
and \(\delta\) are the learning rate and discount factor respectively. \(\tau\) is the rate of update for the target network. Confidence threshold is the value that determines if an environment reward is obtained. Harmonic Mean refers to the averaging we perform for combining the action entropy and reward entropy. 

\section{Results and Analysis}

We perform multiple runs of each algorithm and report the average. Each run uses a different seed for the random initialization of the network weights and environment episodes. Using 24 CPU cores, execution times ranged from 5 hours for key-lock to 10 hours for fetch-push to 15 hours for parking environment. All experiments perform the same number of batch updates for the same number of steps taken. 

\subsection{Results for Key-Lock Domain}

\begin{figure*}[htb]
    \centering
    \subfigure[Asymptotic Performance]{\includegraphics[width=0.66\columnwidth]{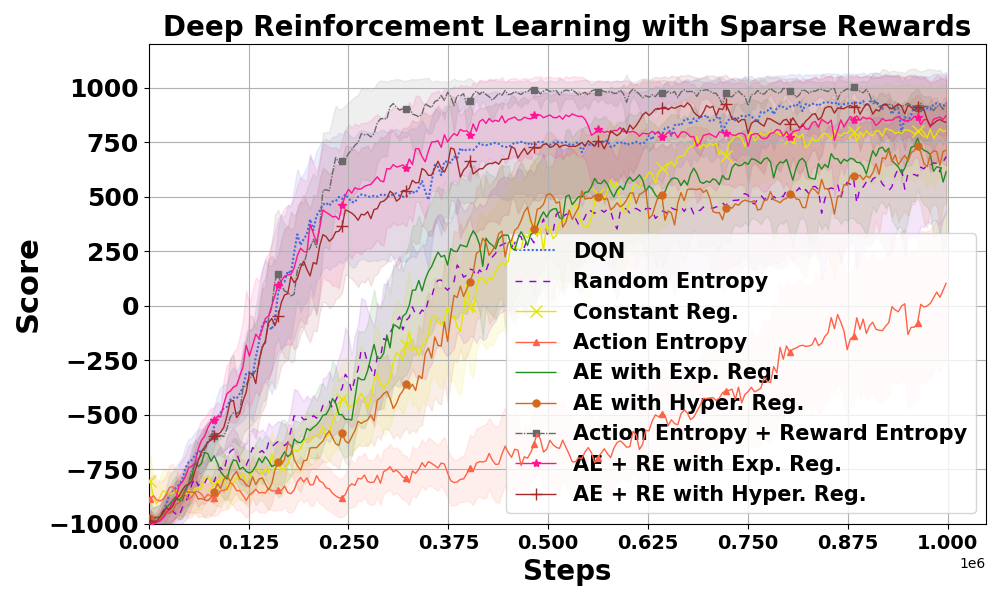}\label{fig:fig1}}
    \subfigure[Rewards Required to Learn]{\includegraphics[width=0.66\columnwidth]{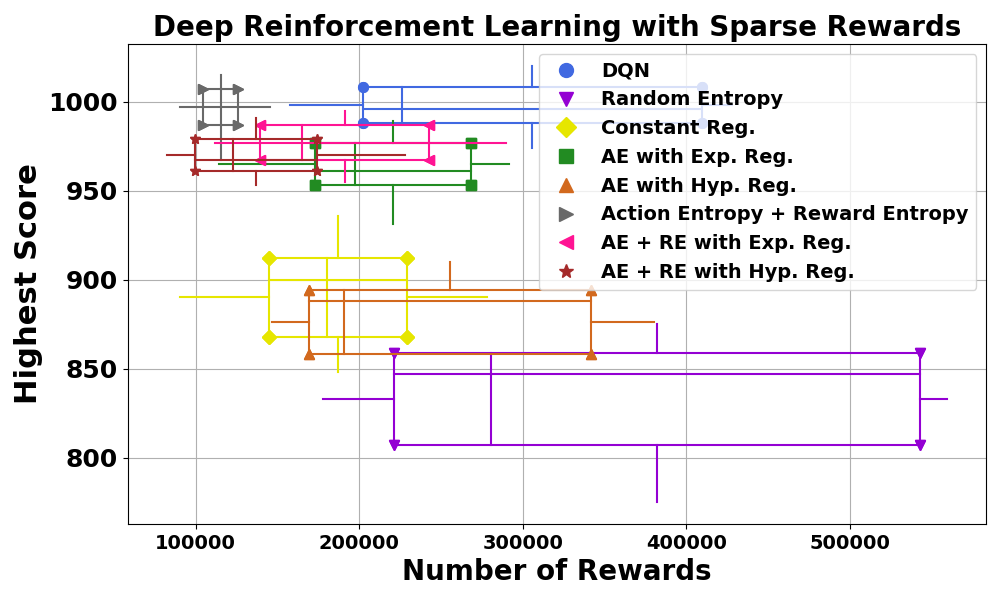}\label{fig:fig2}}
    \subfigure[Model Size vs.\ Performance]{\includegraphics[width=0.66\columnwidth]{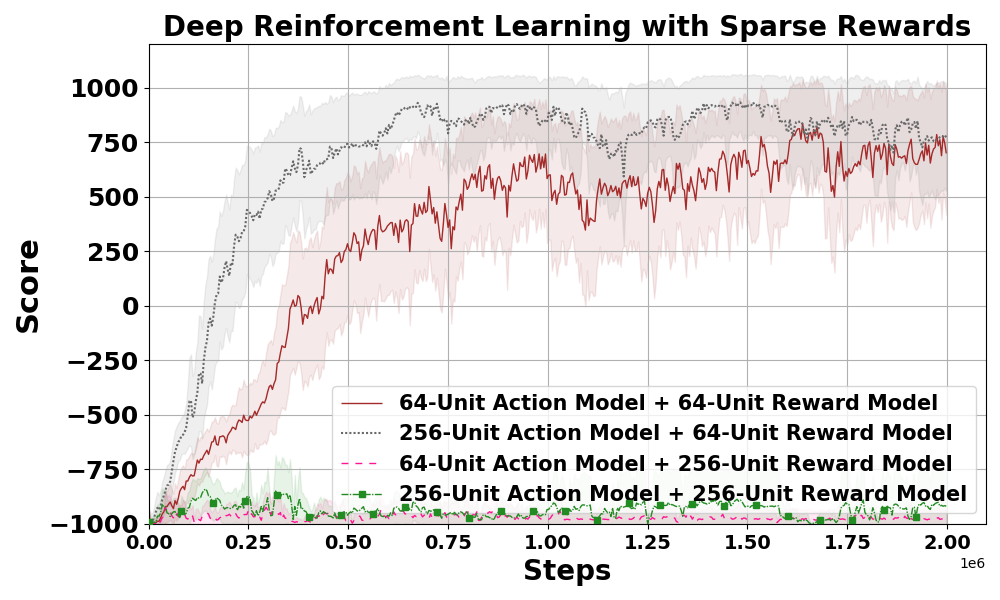}\label{fig:fig3}}
    \caption{Performance of Our Feedback Diminution Algorithms in Key-Lock Domain.}
    \label{fig:fig10}
\end{figure*}

Figure~\ref{fig:fig1} contains the number of training steps across 25 runs for our action entropy and reward entropy algorithms in relation to DQN and random feedback diminution baselines in the key-lock domain.
Constant reg.\ is the baseline reward diminution algorithm applying exponential regularization to a constant entropy value of 1. The random entropy uniformly samples an entropy value between 0 and 1. 
Figure~\ref{fig:fig2} is a box plot representing the number of rewards the agent required to converge to the highest score. The highest score in this case is calculated as the average score of the first five episodes that reach a performance within 5\% of the highest possible score in the environment. The vertical box edges represent the 25th and 75th quantiles for the highest score while the horizontal box edges denote the 25th and 75th quantiles for the number of rewards required to converge. Similarly, the whiskers display the minimum and maximum values excluding outliers. The horizontal and vertical lines within the box represent the median highest score and the median number of rewards required to converge, respectively. The farther the box is to the upper left corner of the plot the better the performance. Figure~\ref{fig:fig3} compares the performance of AE + RE with Hyperbolic Regularization on different sizes of neural networks to get an idea as to how the performance of the reward model changes depending on the neural network architecture. Finally, Table~\ref{table:keyLock} shows the summary statistics for all algorithms. The median score and the rewards required to converge refer to the highest score of the median run and the number of rewards needed to learn the task by the median run respectively, which is equivalent to the vertical median lines of the box plot.

AE + RE reaches the highest score during training while requiring slightly more than 100,000 rewards to converge (Figures~\ref{fig:fig1}~\&~\ref{fig:fig2}), making it the best performing algorithm in this domain. AE + RE with hyperbolic regularization reaches a similar highest score as AE + RE while requiring around 123,000 rewards to converge, making it the second best performing method for this domain (Figure~\ref{fig:fig2}~\&~Table~\ref{table:keyLock}). However, the asymptotic performance of AE + RE with hyperbolic regularization falls slightly behind AE + RE due to one outlier run dragging the average down for AE + RE with hyperbolic regularization (Figure~\ref{fig:fig1}). The DQN baseline requires the highest number of rewards to converge while getting relatively similar performance to AE + RE with exponential regularization (Figure~\ref{fig:fig2}). 

\begin{table}[htb]
    \caption{Summary Results for Key-Lock Environment.}
    \label{table:keyLock}
    \centering
    \footnotesize
     \begin{tabular}{ c c c } 
        \toprule
    \textbf{Algorithm} & \textbf{Median Score} & \textbf{ \# of Rewards} \\ [0.5ex] 
        \midrule
     DQN & 996 &  226,000 \\
    
      Random Entropy & 847 &  280,000 \\
    
      Constant Reg. & 900 &  180,000 \\
    
     Action Entropy & 659 & 233,000  \\
    
    AE with Exp. Reg. & 961 & 197,000 \\ 
     
    AE with Hyper. Reg. & 888 & 191,000 \\
    
    AE + RE & \textbf{997} & \textbf{116,000} \\ 
    
    AE + RE with Exp. Reg. & 977 & 165,000 \\ 
    
    AE + RE with Hyper. Reg & 967 & 123,000 \\ 
        \bottomrule
    \end{tabular}
\end{table}
The constant reg.\ baseline, despite requiring fewer rewards to learn compared to DQN, does not manage to reduce the number of rewards as much as AE + RE with hyperbolic regularization, showing that the entropy selection method is necessary to provide better reduction in external rewards (Figures~\ref{fig:fig1}). AE performs poorly due to not having access to the confidence of the reward model or any type of regularization and gets stuck at a local optima (Figures~\ref{fig:fig1}).
Using a large DQN with a small reward model results in the best performance while the large reward models fail to learn the task (Figures~\ref{fig:fig3}). The large reward models seem to be overfitting to the earlier rewards, making the model overly confident of its policy and not asking for enough environment rewards. Our tuning on the model size indicates it is necessary to have a reward model that is smaller in size to the action model to obtain feedback diminution benefits.

\subsection{Results for Fetch-Push Domain}

\begin{figure}[htb]
    \centering
    \subfigure[Success Rate]{\includegraphics[width=0.99\columnwidth]{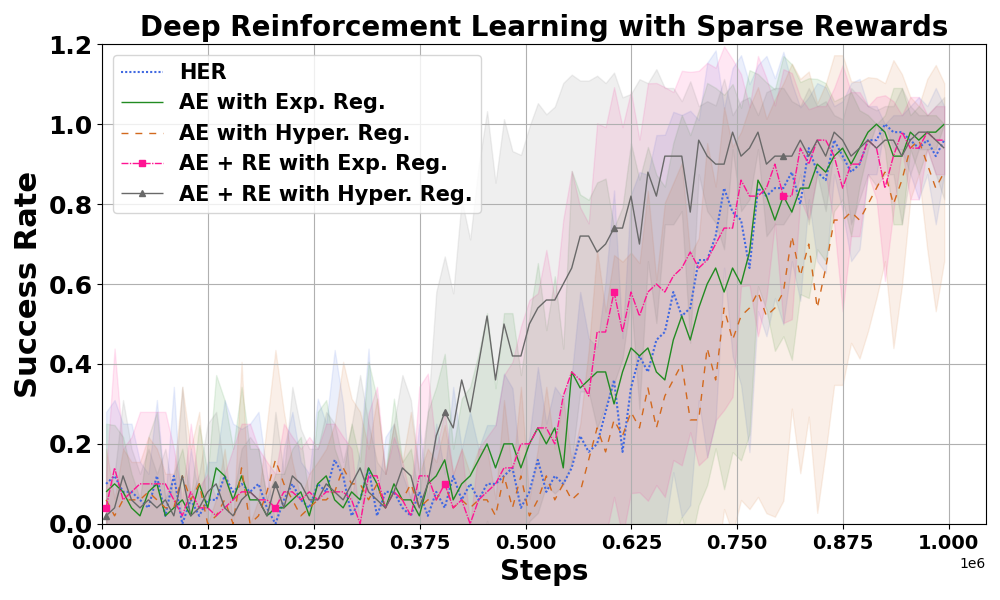}\label{fig:fig5}}
    \subfigure[Rewards Required to Learn]{\includegraphics[width=0.99\columnwidth]{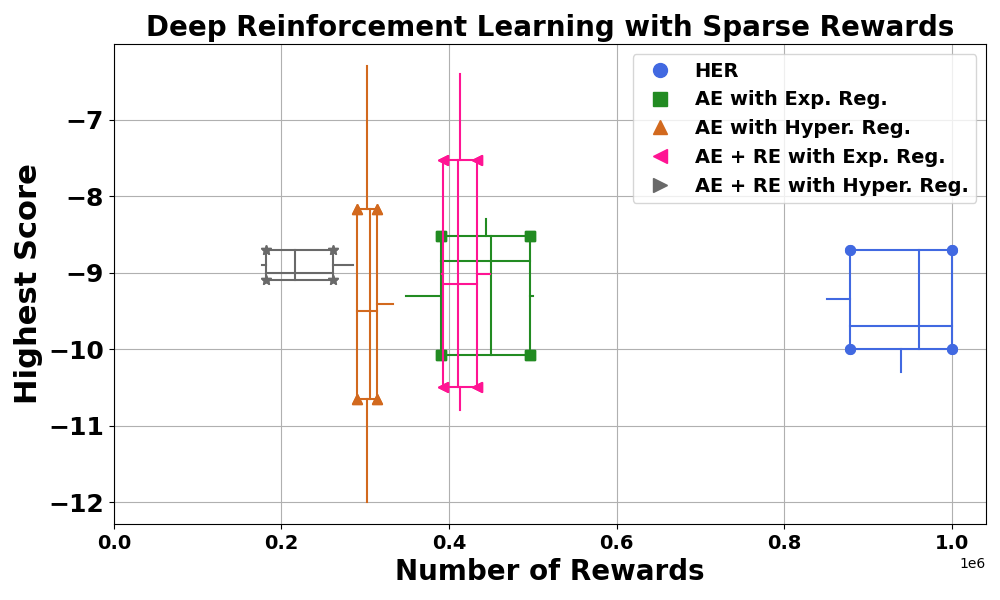}\label{fig:fig6}}
    \caption{Performance in the Fetch-Push Domain.}
    \label{fig:fig20}
\end{figure}

We perform 5 runs of each algorithm in the fetch-push domain and report the average. We use the best performing neural network architecture from (Figures~\ref{fig:fig3}). The confidence intervals appear larger in this domain mainly because of the reduction in the number of runs we execute. Figure~\ref{fig:fig5} shows the asymptotic performance of our algorithms excluding AE and AE + RE. Similar to the key-lock domain, AE performs poorly and fails to reach a high success rate even after 1 million training steps. To make the graphs more readable, we do not show AE or AE + RE as they remain below 0.2 success rate throughout the training. Figure~\ref{fig:fig6} is the box plot for the number of rewards required to convergence.

AE + RE with hyperbolic regularization shows the best asymptotic performance, obtains the highest score, and requires the fewest external rewards, making it the best performing algorithm in this domain (Figures~\ref{fig:fig5}~\&~\ref{fig:fig6}). It also shows the results we obtain in two different environments are consistent with one another since the same algorithm results in near best performance in both domains.  
\begin{table}[htb]
    \caption{Summary Results for Fetch-Push Environment.}
    \label{table:push}
    \centering
    \footnotesize
     \begin{tabular}{ c c c } 
        \toprule
    \textbf{Algorithm} & \textbf{Median Score} & \textbf{\# of Rewards} \\ [0.5ex] 
        \midrule
    HER & -9.7 &  961,000 \\
    
    AE Exp. Reg. & \textbf{-8.85} & 450,000 \\ 
     
    AE Hyper. Reg. & -9.5 & 306,000 \\
    
    AE + RE Exp. Reg. & -9.15 & 411,000 \\ 
    
    AE + RE Hyper. Reg. & -9.0 & \textbf{216,000 }\\ 
        \bottomrule
    \end{tabular}
\end{table}
All the other algorithms have similar performance to the HER baseline in terms of return but they still reduce the number of external rewards by at least 50\% (Figure~\ref{fig:fig6}). AE with exponential regularization obtains the highest median score although it is only slightly better than the variants of AE + RE and the difference could be attributed to the variance in the execution (Table~\ref{table:push}).

\subsection{Results for Parking Domain}

\begin{figure}[htb]
    \centering
    \subfigure[Asymptotic Performance]{\includegraphics[width=0.99\columnwidth]{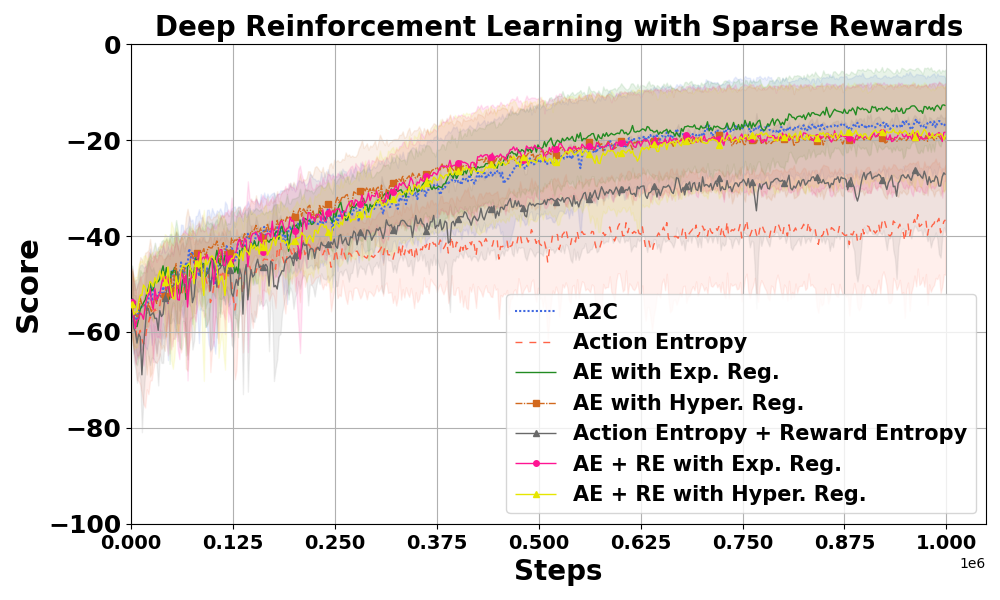}\label{fig:fig7}}
    \subfigure[Rewards Required to Learn]{\includegraphics[width=0.99\columnwidth]{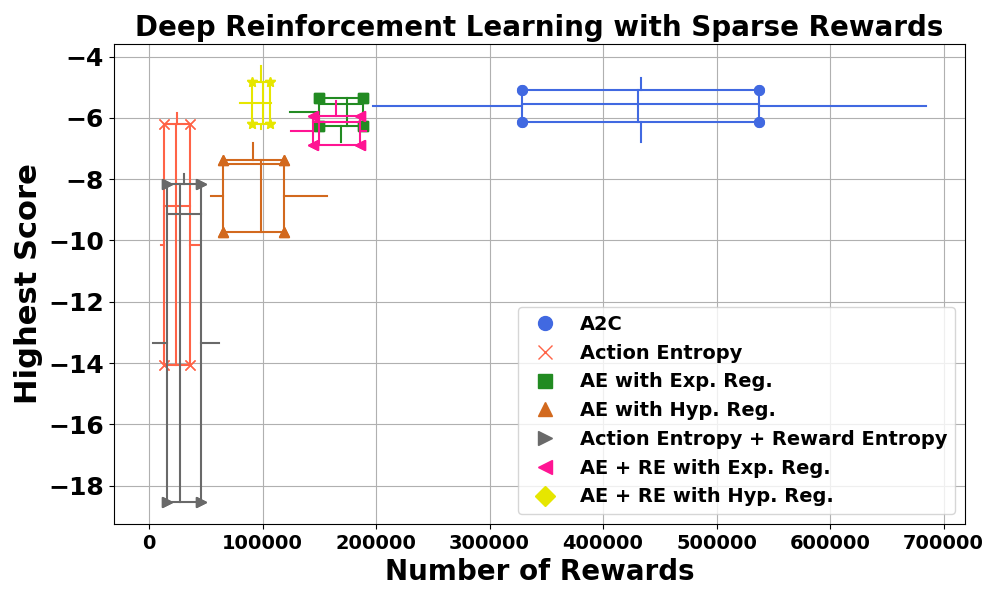}\label{fig:fig8}}
    \caption{Performance in the Parking Domain.}
    \label{fig:fig30}
\end{figure}

We report the average of 25 runs in the parking domain, using the best performing neural network architecture from (Figure~\ref{fig:fig3}). Figure~\ref{fig:fig7} shows the asymptotic performance against advantage actor-critic (A2C). Figure~\ref{fig:fig8} is the box plots for the number of rewards required to converge. 

AE with exponential reg.\ gives the best average score (although it is only slightly better than the A2C baseline) (Figure~\ref{fig:fig7}), but requires more environment rewards to learn the task compared to the variants of AE + RE (Figure~\ref{fig:fig8}). AE + RE with hyperbolic regularization produces the highest score while needing a very low number of environment rewards to converge (Figure~\ref{fig:fig8}), arguably making it the best performing algorithm again. AE + RE and AE manage to reduce the number of rewards needed to converge at the expense of performance. They show poor asymptotic performance similar to prior domains (Figure~\ref{fig:fig7}) and show a large variation in the highest score they obtain (Figure~\ref{fig:fig8}). AE only requires 24,000 rewards to learn, providing close to 95\% improvement and suggesting that denser external rewards might be more prone to reward skipping by the agent (Table~\ref{table:park}). The baseline algorithm obtains a similar highest score as the regularized AE + RE but it requires more than four times the number of environment rewards to converge (Figure~\ref{fig:fig8}). The performance of AE + RE with hyperbolic regularization and AE with exponential reg.\ remain consistent with the other domains despite the parking environment having a denser external reward function. This further supports reducing the need for environment rewards across multiple network architectures over three domains. 

We also see that some of the feedback diminution algorithms reach higher scores than the baselines having access to true rewards. Only the feedback diminution algorithms making use of hyperbolic or exponential regularization manage to outperform the baseline algorithms in some domains, which makes us reason that the effect of the regularization allows the algorithm to avoid suboptimal policies by providing interleaved feedback which results in the feedback diminution method outperforming the baselines.

\begin{table}[htb]
    \caption{Summary Results for Parking Environment.}
    \label{table:park}
    \centering
    \footnotesize
     \begin{tabular}{ c c c } 
        \toprule
    \textbf{Algorithm} & \textbf{Median Score} & \textbf{\# of Rewards} \\ [0.5ex] 
        \midrule
     A2C & -5.53 &  431,000 \\
    
     AE & -8.88 & \textbf{24,000}  \\
    
    AE Exp. Reg. & -5.55 & 174,000 \\ 
     
    AE Hyper. Reg. & -7.50 & 99,000 \\
    
    AE + RE & -9.15 & 27,000 \\ 
    
    AE + RE Exp. Reg. & -6.13 & 150,000 \\ 
    
    AE + RE Hyper. Reg. & \textbf{-5.51} & 101,000 \\ 
        \bottomrule
    \end{tabular}
\end{table}

\subsection{Choosing the Best Performing Approach}

Having a feedback diminution method that can provide consistent improvements across multiple domains is crucial since it would be impractical to try all regularization methods and entropy metrics every time the algorithm is adapted to a new domain or framework. Due to the variations in performance of the feedback diminution algorithms on different domains, there emerges the need for selecting the proper approach that would be suitable for most domains and real-world applications. If we exclude the non-regularized methods due to their poor scores in some environments, we see that AE + RE with hyperbolic regularization produces the best performance in the key-lock and robotics domains in terms of the rewards required to converge while being close to the best performing algorithm in the parking domain. If the purpose is to reduce the feedback dependence while not compromising the performance of the RL algorithm, AE + RE with hyperbolic regularization presents the best option for applying feedback diminution in RL frameworks. 

\section{Conclusion}

We presented a novel approach to reducing the need to sample environment rewards using entropy as a heuristic. We detailed two versions of the approach, one using only the entropy of the action model
 and another that incorporates both action and reward entropies. We adapted variants of DQN, A2C, and HER algorithms to
compare against our algorithm and evaluated performance on the key-lock, robotics, and parking domains using discrete and continuous action spaces. Our results show that our approach was able to match the performance of a randomly generated
entropy baseline, a constant regularization baseline, and the comparison criteria while requiring fewer environmental rewards to learn in all cases. 
In the future, we intend to analyze in what parts of the state space the agent is more likely to request rewards to obtain a better understanding of how the characteristics of the environment affect the feedback diminution process.

\bibliography{ref}

\end{document}